\documentclass[runningheads]{llncs}
\usepackage{graphicx}
\usepackage{amsmath,amssymb} 
\usepackage{color}
\usepackage[width=122mm,left=12mm,paperwidth=146mm,height=193mm,top=12mm,paperheight=217mm]{geometry}

\usepackage{hhline}
\usepackage{multirow}
\usepackage[pagebackref=false,breaklinks=true,letterpaper=true,colorlinks,bookmarks=false]{hyperref}

\newcommand{\eg} {\emph{e.g. }}

\newcommand{\ie} {\emph{i.e. }}

\begin{document}
\title{Two at Once: Enhancing Learning and Generalization Capacities via IBN-Net
} 

\titlerunning{IBN-Net}
%
\author{Xingang Pan\inst{1} \and
Ping Luo\inst{1} \and
Jianping Shi\inst{2} \and
Xiaoou Tang\inst{1}}
%
\authorrunning{X. Pan, P. Luo, J. Shi, and X. Tang}
%

\institute{CUHK-SenseTime Joint Lab, The Chinese University of Hong Kong \\
\email{\{px117,pluo,xtang\}@ie.cuhk.edu.hk}\\
\and SenseTime Group Limited \\
\email{shijianping@sensetime.com}}
\maketitle              
\setcounter{footnote}{0}
\begin{abstract}
Convolutional neural networks (CNNs) have achieved great successes in many computer vision problems.
Unlike existing works that designed CNN architectures to improve performance on a single task of a single domain and not generalizable, 
we present IBN-Net, a novel convolutional architecture, which remarkably enhances a CNN's modeling ability on one domain (\eg Cityscapes) as well as its generalization capacity on another domain (\eg GTA5) without finetuning. 
IBN-Net carefully integrates Instance Normalization (IN) and Batch Normalization (BN) as building blocks, and
%
can be wrapped into many advanced deep networks to improve their performances.
This work has three key \emph{contributions}.
(1) By delving into IN and BN, we disclose that IN learns features that are invariant to appearance changes, such as colors, styles, and virtuality/reality, 
while BN is essential for preserving content related information.
(2) IBN-Net can be applied to many advanced deep architectures, such as DenseNet, ResNet, ResNeXt, and SENet, and consistently improve their performance without increasing computational cost.
\footnote{Code and models are available at \textit{https://github.com/XingangPan/IBN-Net}}
(3)
When applying the trained networks to new domains, \eg from GTA5 to Cityscapes,
IBN-Net achieves comparable improvements as domain adaptation methods, even without using data from the target domain.
With IBN-Net, we won the 1st place on the WAD 2018 Challenge Drivable Area track, with an mIoU of 86.18\%.

\end{abstract}
\section{Introduction}

\begin{figure}[t]
\centering
\includegraphics[width=9.5cm,height=2.3in]{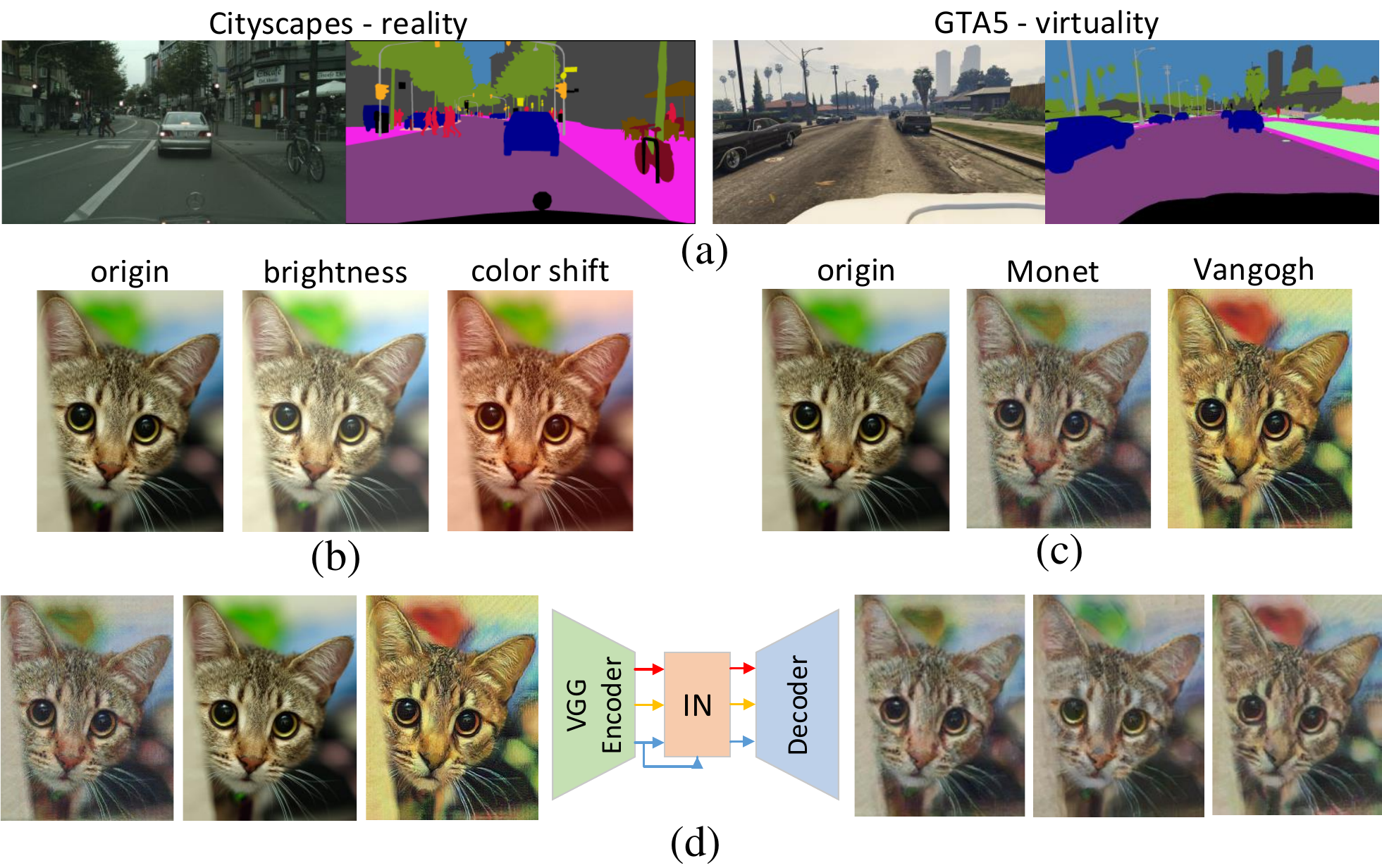}
\caption{(a) visualizes two example images (left) and their segmentation maps (right) selected from Cityscapes \cite{cordts2016cityscapes} and GTA5 \cite{Richter_2016_ECCV} respectively. These samples have similar categories and scene configurations when comparing their segmentation maps, but their images are from different domains, \ie reality and virtuality. (b) shows simple appearance variations, while those of complex appearance variations are provided in (c). (d) proves that Instance Normalization (IN) is able to filter out complex appearance variance. The style transfer network used here is AdaIN \cite{huang2017arbitrary}. 
(Best viewed in color) }
\label{fig:intro}
\end{figure}

Deep convolutional neural networks (CNNs) have improved performance of many tasks in computer vision, such as image recognition \cite{krizhevsky2012imagenet}, object detection \cite{ren2015faster}, and semantic segmentation \cite{chen2017deeplab}.
However, existing works mainly design network architectures to solve the above problems on a single domain, for example, improving scene parsing on the real images of Cityscape dataset \cite{cordts2016cityscapes,pan2018SCNN}.
When these networks are applied to the other domain of this scene parsing task, such as the virtual images of GTA5 dataset \cite{Richter_2016_ECCV}, their performance would drop notably.
This is due to the appearance gap between the images of these two datasets, as shown in Fig.\ref{fig:intro} (a).

A natural solution to solve the appearance gap is by using transfer learning. For instance, by finetuning a CNN pretrained on Cityscapes using the data from GTA5, we are able to adapt the features learned from Cityscapes to GTA5, where accuracy can be increased.
But even so, the appearance gap is not eliminated, because when applying the finetuned CNN back to Cityscapes, the accuracy would be significantly degraded.
How to address large diversity of appearances by designing deep architectures?
It is a key challenge in computer vision.

The answer is to induce appearance invariance into CNNs.
This solution is obvious but non-trivial.
For example, there are many ways to produce the property of spatial invariance in deep networks, such as max pooling~\cite{krizhevsky2012imagenet}, deformable convolution~\cite{dai2017deformable}, which are invariant to spatial variations like poses, viewpoints, and scales, but are not invariant to variations of image appearances.
As shown in Fig.\ref{fig:intro} (b), when the appearance variance of two datasets are simple and known beforehand, such as lightings and infrared, they can be reduced by explicitly augmenting data.
However, as shown in Fig.\ref{fig:intro} (c), when appearance variance are complex and unknown, such as arbitrary image styles and virtuality, the CNNs have to learn to reduce them by introducing new component into their deep architectures.

To this end, we present IBN-Net, a novel convolutional architecture, which learns to \emph{capture} and \emph{eliminate} appearance variance, while \emph{maintains} discrimination of the learned features.
IBN-Net carefully integrates Instance Normalization (IN) and Batch Normalization (BN) as building blocks, enhancing both its learning and generalization capacity.
It has two appealing benefits that previous deep architectures do not have.

First, different from previous CNN structures that isolate IN and BN, IBN-Net unifies them by delving into their learned features.
For example, many recent advanced deep architectures employed BN as a key component to improve their learning capacity in high-level vision tasks such as image recognition \cite{he2016deep,xie2017aggregated,hu2017squeeze,huang2017densely}, while IN was often combined with CNNs to remove variance of images on low-level vision tasks such as image style transfer \cite{ulyanov2017improved,dumoulin2016learned,huang2017arbitrary}.
But the different characteristics of their learned features and the impact of their combination have not been disclosed in existing works.
In contrast, IBN-Net shows that combining them in an appropriate manner improves both learning and generalization capacities.

\begin{figure*}[t]
\centering
\includegraphics[width=7.2cm]{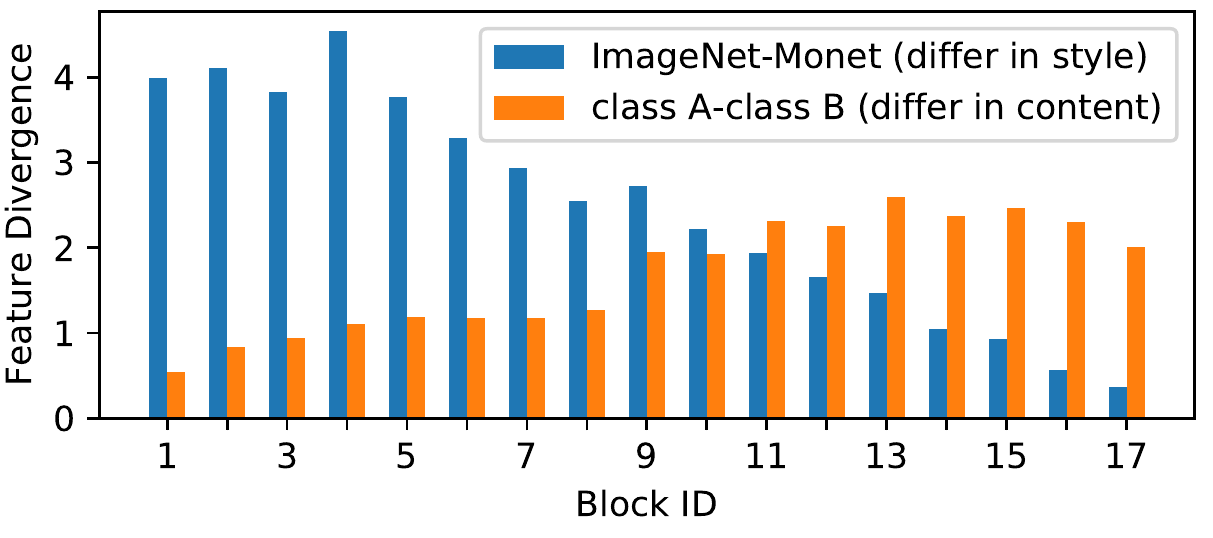}
\caption{\label{casestudy} (a) Feature divergence calculated from image sets with appearance difference (blue) and content difference (orange). We show the results of the 17 features after the residual blocks of ResNet50. The detailed definition of feature divergence is given in Section 4.3. The orange bars are enlarged 10 times for better visualization.}

\end{figure*}

Second, our IBN-Net keeps IN and BN features in shallow layer and BN features in higher layer, inheriting from the statistical merit of feature divergence under different depth of a network. 
As shown in Fig.\ref{casestudy}, the x-axis denotes the depth of a network and the y-axis shows feature divergence calculated via symmetric KL divergence. 
When analyzing the depth-vs-divergence in ImageNet original with its Monet version (blue bars), the divergence decreases as layer depth increases, manifesting the appearance difference mainly lies in shallow layers. 
On the contrary, compared with two disjoint ImageNet splits (orange bar), the object level difference attributes to majorly higher layer divergence and partially low layer ones.
Based on these observations, we introduce IN layers to CNNs following two rules. 
Firstly, to reduce feature variance caused by appearance in shallow layers while not interfering the content discrimination in deep layers, we only add IN layers to the shallow half of the CNNs. 
Secondly, to also preserve image content information in shallow layers, we replace the original BN layers to IN for a half of features and BN for the other half. 
These give rise to our IBN-Net.

Our \textbf{contributions} can be summarized as follows:

(1) A novel deep structure, IBN-Net, is proposed to \emph{improve both} learning and generalization capacities of deep networks.
For example, IBN-Net50 achieves 22.54\%/6.32\% and 51.57\%/27.15\% top1/top5 errors on the original validation set of ImageNet~\cite{deng2009imagenet} and a new validation set after style transformation respectively, outperforming ResNet50 by 1.73\%/0.76\% and 2.17\%/2.94\%, when they have similar numbers of parameters and computational cost.

(2) By delving into IN and BN, we disclose the key characteristics of their learned features, where IN provides visual and appearance \emph{invariance}, while BN accelerates training and preserves \emph{discriminative} features.
This finding is important to understand them, and helpful to design the architecture of IBN-Net, where IN is preferred in shallow layers to remove appearance variations, whereas its strength in deep layers should be reduced in order to maintain discrimination.
The component of IBN-Net can be used to re-develop many recent deep architectures, improving both their learning and generalization capacities, but keeping their computational cost unchanged.
For example, by using IBN-Net, DenseNet169 \cite{huang2017densely}, ResNet101 \cite{he2016deep}, ResNeXt101 \cite{xie2017aggregated}, and SE-ResNet101 \cite{hu2017squeeze}, outperform their original versions by 0.79\%, 1.09\%, 0.43\%, and 0.43\% on ImageNet respectively.
These re-developed networks can be utilized as \emph{strong backbones} in many tasks in future researches. 

(3) IBN-Net significantly improves performance across domains. 
By taking scene understanding as an example under a cross-evaluation setting, \ie training a CNN on Cityscapes and evaluating it on GTA5 without finetuning and vice versa, ResNet50 integrated with IBN-Net improves its counterpart by 8.5\% and 7.5\% respectively.
It also notably reduces sample size when finetuning GTA5 pretrained model on Cityscapes.
For instance, it achieves a segmentation accuracy of 65.5\% when finetuning using just 30\% training data from Cityscapes, compared to 63.8\% of ResNet50 alone, which is finetuned using all training data.

\section{Related Works}

The previous work related to IBN-Net are described in three aspects, including invariance of CNNs, network architectures, and methods of domain adaptation and generalization.

\textbf{Invariance in CNNs}.
Several modules \cite{krizhevsky2012imagenet,dai2017deformable,srivastava2014dropout,ulyanov2017improved,ioffe2015batch} have been proposed to improve a CNN's modeling capacity, or reduce overfitting to enhance its generalization capacity on a single domain.
These methods typically achieved the above purposes by introducing specific kinds of invariance into the architectures of CNNs.
For example, max pooling~\cite{krizhevsky2012imagenet} and deformable convolution~\cite{dai2017deformable} introduce spatial invariance to CNNs, thus increasing their robustness to spatial variations such as affine, distortion, and viewpoint transformations.
And dropout~\cite{srivastava2014dropout} and batch normalization (BN)~\cite{ioffe2015batch} can be treated as regularizers to reduce the effects of sample noise in training.
When image appearances are presented, simple appearance variations such as color or brightness shift could simply be eliminated by normalizing each RGB channel of an image with its mean and standard deviation.
For more complex appearance transforms such as style transformations, recent studies have found that such information could be encoded in the mean and variance of the hidden feature maps~\cite{dumoulin2016learned,huang2017arbitrary}.
Therefore, the instance normalization (IN)~\cite{ulyanov2017improved} layer shows potential to eliminate such appearance differences.

\textbf{CNN Architectures}.
Since CNNs have shown compelling modeling capacity over traditional methods, their architectures have gone through a number of developments.
Among them one of the most widely used is the residual network (ResNet)~\cite{he2016deep}, which uses short cut to alleviate training difficulties of very deep networks.
Since then a number of variants of ResNet were proposed.
Compared to ResNet, ResNeXt~\cite{xie2017aggregated} improves modeling capacity by increasing `cardinality' of ResNet. It is implemented by using group convolutions.
In practice, increasing cardinality increases runtime in modern deep learning frameworks.
Moreover, squeeze-and-excitation network (SENet)~\cite{hu2017squeeze} introduces channel wise attention into ResNet.
It achieves better performance on ImageNet compared to ResNet, but it also increases number of network parameters and computations.
The recently proposed densely connected networks (DenseNet)~\cite{huang2017densely} uses concatenation to replace short-cut connections.
It was proved to be more efficient than ResNet.

However, there are two limitations in the above CNN architectures.
Firstly, the limited basic modules prevent them from gaining more appealing properties.
For example, all these architectures are simply composed of convolutions, BNs, ReLUs, and poolings.
The only difference among them is how these modules are organized.
However, the composition of these layers are naturally vulnerable by appearance variations.
Secondly, the design goal of these models is to achieve strong modeling capacity on a single task of a single domain, while their capacities to generalize to new domains are still limited.

In the field of image style transfer, some methods employ IN to help remove image contrast~\cite{ulyanov2017improved,dumoulin2016learned,huang2017arbitrary}.
Basically, this helps the models transfer images to different styles.
However, the invariance property of image appearance has not been successfully introduced to aforementioned CNNs, especially in high-level tasks such as image classification or semantic segmentation. 
This is because IN drops useful content information presented in the hidden features, impeding modeling capacity as proved in \cite{ulyanov2017improved}.

\textbf{Improve Performances across Domains}.
Alleviating the drop of performances caused by appearance gap between different domains is an important problem.
One natural approach is to use transfer learning such as finetuning the model on the target domain.
However, this requires human annotations of the target domain, and the performances of the finetuned models would then drop when they are applied on the source domain.
There are a number of domain adaptation approaches which use the statistics of the target domain to facilitate adaptation.
Most of these works address the problem by reducing feature divergences between two domains through carefully designed loss functions, like maximum mean discrepancy (MMD) ~\cite{tzeng2014deep,long2015learning}, correlation alignment (CORAL)~\cite{sun2016deep}, and adversarial loss~\cite{tzeng2017adversarial,hoffman2016fcns}.
Besides, \cite{sankaranarayanan2017unsupervised} and~\cite{hoffman2017cycada} use generative adversarial networks (GAN) to transfer images between two domains to help adaptation, but required independent models for the two domains.
AdaBN~\cite{li2016revisiting} provides a simple approach for domain adaptation simply by adjusting the statistics of all BN layers using data from the target domain.
Our method does not rely on any specific target domain and there is no need to adjust any parameters.
There are two main limitations in transfer learning and domain adaptation.
First, in real applications it is expensive and difficult to collect data that covers all possible scenarios in the target domain.
Second, most state-of-the-art methods employ different model weights for the source and target domains in order to improve performance.
But the ideal case is that one model could adapt to all domains.

Another paradigm towards this problem is domain generalization, which aims to acquire knowledge from a number of related source domains and apply it to a new target domain whose statistics is unknown during training.
Existing methods typically design algorithms to learn domain agnostic representations or design models that capture common aspects from the domains, such as~\cite{khosla2012undoing}~\cite{muandet2013domain}~\cite{ghifary2015domain}.
However, for real applications it is often hard to acquire data from a number of related source domains, and the performance highly depends on the series of source domains.

In this work, we increase the modeling capacity and generalization ability across domains by designing a new CNN architecture, IBN-Net.
The benefit is that we do not require either target domain data or related source domains, unlike existing domain adaptation and generalization methods.
The improvement of generalization across domains is achieved by designing architectures with built-in appearance invariance.
Our method is extremely useful for the situations that the target domain data are unobtainable, where traditional domain adaptation cannot be applied.
For more detailed comparison of our method with related works, please refer to our supplementary material.

\section{Method}

\subsection{Background}
\textbf{Batch normalization}~\cite{ioffe2015batch} enables larger learning rate and faster convergence by reducing the internal covariate shift during training CNNs.
It uses the mean and variance of a mini-batch to normalize each feature channels during training,
while in inference phase, BN uses the global statistics to normalize features.
Experiments have shown that BN significantly accelerates training, and could improve the final performance meanwhile.
It has become a standard component in most prevalent CNN architectures like Inception~\cite{szegedy2015going}, ResNet~\cite{he2016deep}, DenseNet~\cite{huang2017densely}, etc.

Unlike batch normalization, \textbf{instance normalization}~\cite{ulyanov2017improved} uses the statistics of an individual sample instead of mini-batch to normalize features.
Another important difference between IN and BN is that IN applies the same normalize procedure for both training and inference.
Instance normalization has been mainly used in the style transfer field~\cite{ulyanov2017improved,dumoulin2016learned,huang2017arbitrary}.
The reason for IN's success in style transfer and similar tasks is that, these tasks try to change image appearance while preserving content, and IN allows to filter out instance-specific contrast information from the content.
Despite these successes, IN has not shown benefits for high-level vision tasks like image classification and semantic segmentation.
Ulyanov \textit{et al}~\cite{ulyanov2017improved} have given primary attempt adopting IN for image classification, but got worse results than CNNs with BN.

In a word, batch normalization preserves discrimination between individual samples, but also makes CNNs vulnerable to appearance transforms.
And instance normalization eliminates individual contrast, but diminishes useful information at the same time. 
Both methods have their limitations.
In order to introduce appearance invariance to CNNs without hurting feature discrimination, here we carefully unify them in a single deep hierarchy.

\subsection{Instance-Batch Normalization Networks}

\begin{figure*}[!t]
\centering
\includegraphics[width=9.5cm]{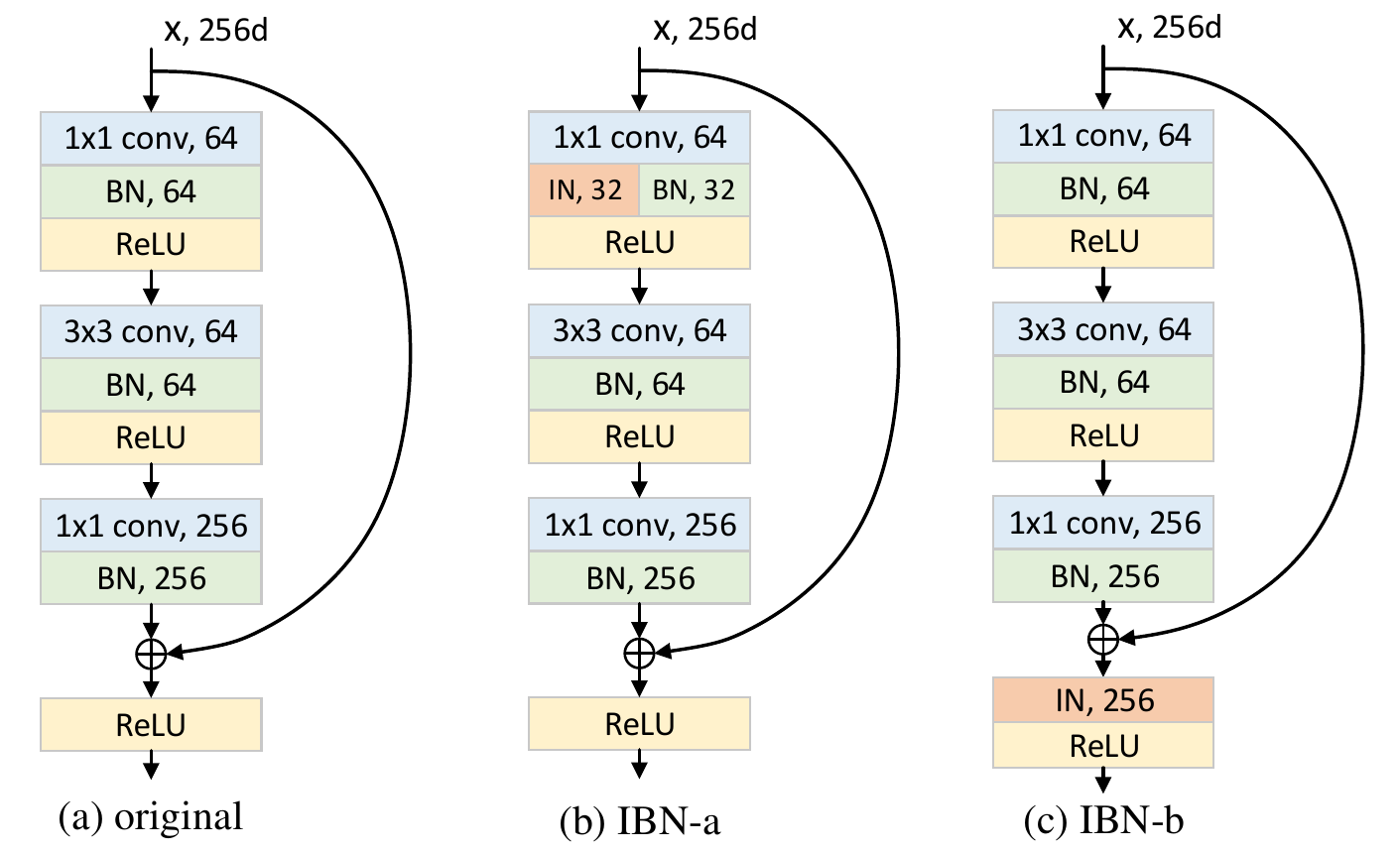}
\caption{\label{BINNet} Instance-batch normalization (IBN) block. }
\end{figure*}

Our architecture design is based on an important observation: as shown in Fig.~\ref{casestudy}(a)(b), for BN based CNNs, the feature divergence caused by appearance variance mainly lies in shallow half of the CNN, while the feature discrimination for content is high in deep layers, but also exists in shallow layers.
Therefore we introduce INs following two rules.
Firstly, in order not to diminish the content discrimination in deep features, we do not add INs in the last part of CNNs.
Secondly, in order to also preserve content information in shallow layers, we keep part of the batch normalized features.

To provide instance for discussion, we describe our method based on the classic residual networks (ResNet).
ResNet mainly consists of four groups of residual blocks, with each block having the structure as shown in Fig.~\ref{BINNet}(a).
Following our first rule, we only add IN to the first three groups (conv2\_x-conv4\_x) and leave the fourth group (conv5\_x) as before.
For a residual block, we apply BN for half channels and IN for the others after the first convolution layer in the residual path, as Fig.~\ref{BINNet}(b) shows.
There are three reasons to do so.
Firstly, as \cite{he2016identity} pointed out, a clean identity path is essential for optimizing ResNet, so we add IN to the residual path instead of identity path.
Secondly, in the residual learning function \(\mathbf{y}=\mathcal{F}(\mathbf{x},\{W_i\})+\mathbf{x}\), the residual function \(\mathcal{F}(\mathbf{x},\{W_i\})\) is learned to align with \(\mathbf{x}\) in the identity path. 
Therefore IN is applied to the first normalization layer instead of the last to avoid misalignment.
Thirdly, the half BN half IN scheme comes from our second design rule as discussed before.
This gives rise to our instance-batch normalization network (IBN-Net).

This design is a pursuit of model capacity. 
On one hand, INs enable the model to learn appearance invariant features so that it could better utilize the images with high appearance diversity within one dataset. 
On the other hand, INs are added in a moderate way so that content related information could be well preserved.
We denote this model as IBN-Net-a.
To take full use of IN's potential for generalization, in this work we also study another version, which is IBN-Net-b.
Since appearance information could be either preserved in residual path or identity path, we add IN right after the addition operation, as shown in Fig.~\ref{BINNet}(c).
To not deteriorate optimization for ResNet, we only add three IN layers after the first convolution layer (conv1) and the first two convolution groups (conv2\_x, conv3\_x).

\textbf{Variants of IBN-Net.}

\begin{figure*}[!t]
\centering
\includegraphics[width=12.5cm]{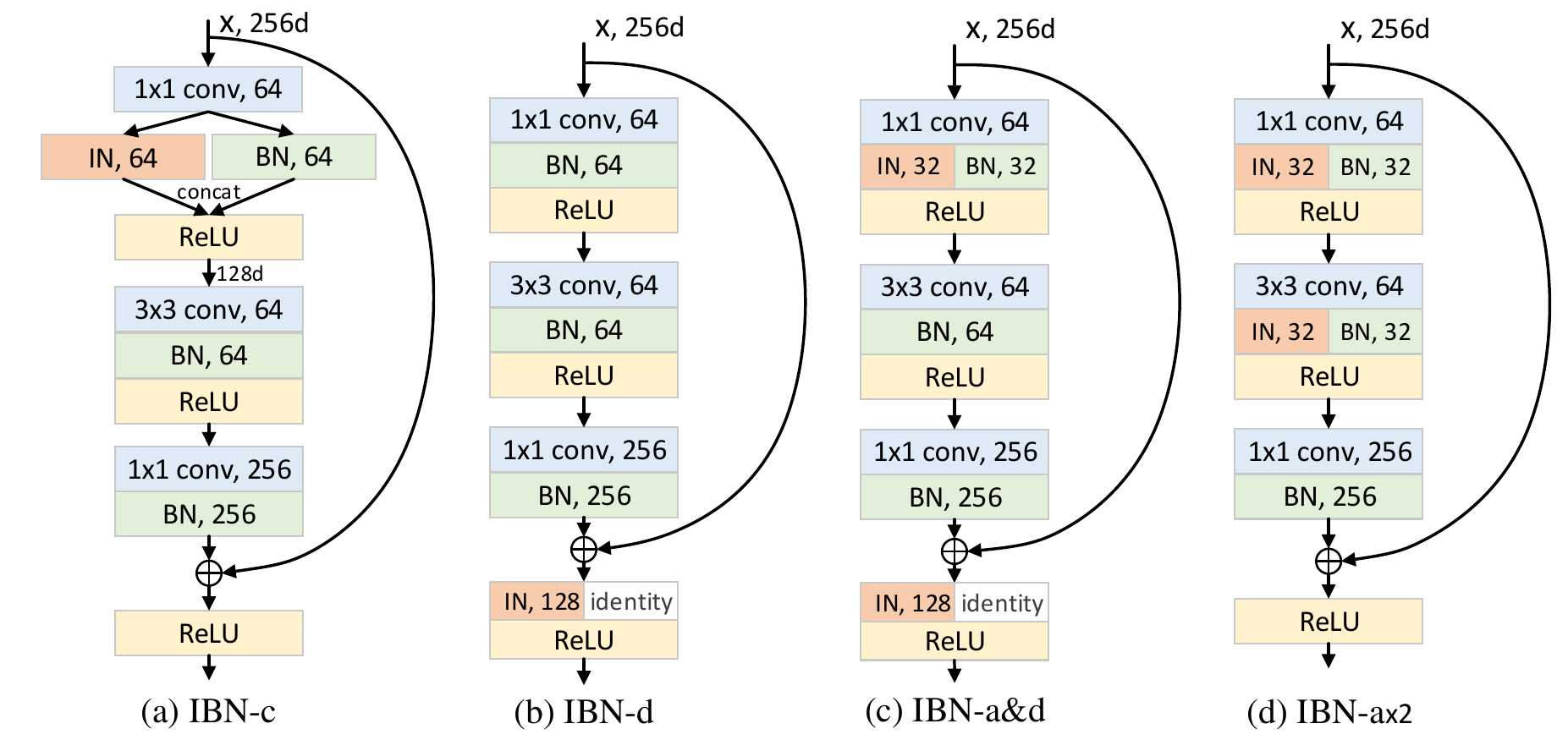}
\caption{\label{BINNet_v} Variants of IBN block. }
\end{figure*}

The two types of IBN-Net described above are not the only ways to utilize IN and BN in CNNs.
In the experiments we will also study some interesting variants, as shown in Fig.~\ref{BINNet_v}.
For example, to keep both generalizable and discriminative features, another natural idea is to feed the feature to both IN and BN layers and then concatenate their outputs, as in Fig.~\ref{BINNet_v}(a), but this would introduce more parameters. 
And the idea of keeping two kind of features also be applied to the IBN-b, giving rise to Fig.~\ref{BINNet_v}(b).
We may also combine these schemes as Fig.~\ref{BINNet_v}(c)(d) do.
Discussions about these variants would be given in the experiments section.

\section{Experiments}

We evaluate IBN-Net on both classification and semantic segmentation tasks on the ImageNet and Cityscapes-GTA5 dataset respectively.
In both tasks, we study our models' modeling capacity within one dataset and their generalization under appearance transforms.

\subsection{ImageNet Classification}

We evaluate our method on the ImageNet~\cite{deng2009imagenet} 2012 classification dataset with 1000 object classes.
It has 1.28 million images for training and 50k images for validation.
Data augmentation includes random scale, random aspect ratio, random crop, and random flip.
We use the same training policy as in~\cite{gross2016training}, and apply \(224\times 224\) center crop during testing.

\textbf{Generalization to Appearance Transforms.}
We first evaluate the models' generalization to many kinds of appearance transforms including shift in color, brightness, contrast, and style transform, which is realized using CycleGAN~\cite{zhu2017unpaired}. 
The models are trained merely on ImageNet training set and evaluated on validation set with the appearance transforms mentioned.
The result for the original ResNet50 and our IBN-Net versions are given in Table.~\ref{generalization}.

\setlength{\tabcolsep}{4pt}
\begin{table}[!t]
\centering
\caption{Results on ImageNet validation set with appearance transforms. The performance drops are given in brackets.}
\label{generalization}
\resizebox{8.0cm}{1.9cm}{
\begin{tabular}{lccc}
\hline
\multirow{2}{*}{\begin{tabular}[c]{@{}l@{}}appearance\\ transform\end{tabular}} & ResNet50 \cite{he2016deep}                                                            & IBN-Net50-a                                                         & IBN-Net50-b                                                                  \\ \cline{2-4} 
                                                                                & top1/top5 err.                                                      & top1/top5 err.                                                      & top1/top5 err.                                                               \\ \hline
origin                                                                          & 24.27/7.08                                                          & \textbf{22.54/6.32}                                                 & 23.64/6.86                                                                   \\ \hline
RGB+50                                                                          & \begin{tabular}[c]{@{}c@{}}28.22/9.64\\ (3.94/2.56)\end{tabular}   & \begin{tabular}[c]{@{}c@{}}25.54/8.03\\ (3.00/1.71)\end{tabular}   & \textbf{\begin{tabular}[c]{@{}c@{}}23.82/6.96\\ (0.18/0.10)\end{tabular}}   \\ \hline
R+50                                                                            & \begin{tabular}[c]{@{}c@{}}27.53/8.78\\ (3.26/1.70)\end{tabular}    & \begin{tabular}[c]{@{}c@{}}25.20/7.56\\ (2.66/1.24)\end{tabular}    & \textbf{\begin{tabular}[c]{@{}c@{}}25.10/7.43\\ (1.46/0.57)\end{tabular}}    \\ \hline
std \(\times 1.5\)                                                         & \begin{tabular}[c]{@{}c@{}}40.01/19.08\\ (15.74/12.00)\end{tabular} & \begin{tabular}[c]{@{}c@{}}35.97/16.22\\ (13.43/9.90)\end{tabular}  & \textbf{\begin{tabular}[c]{@{}c@{}}23.64/6.86\\ (0.00/0.00)\end{tabular}}    \\ \hline 
Monet                                                                           & \begin{tabular}[c]{@{}c@{}}54.51/29.32\\ (30.24/22.24)\end{tabular} & \begin{tabular}[c]{@{}c@{}}51.57/27.15\\ (29.03/20.83)\end{tabular} & \textbf{\begin{tabular}[c]{@{}c@{}}50.45/25.22\\ (26.81/18.36)\end{tabular}} \\ \hline                                            
\end{tabular}
}
\end{table}
\setlength{\tabcolsep}{1.4pt}

From the results we can see that IBN-Net-a achieves both better generalization and stronger capacity.
When applied to images with new appearance domains, it shows less performance drop than the original ResNet.
Meanwhile, its top1/top5 error on the original images is significantly improved by 1.73\%/0.76\%, showing that the model capacity is also improved.
For IBN-Net-b, generalization is significantly enhanced, as the performance drops on new image domains are largely reduced.
This shows that IN does help CNNs to generalize.
Meanwhile, its performance on the original images also increases a little, showing that although IN removed discrepancy of feature mean and variance, content information could be well preserved in the spatial dimension.

\setlength{\tabcolsep}{4pt}
\begin{table}[!t]
\begin{center}
\caption{Results of IBN-Net over other CNNs on ImageNet validation set. The performance gains are shown in the brackets. More detailed descriptions of these IBN-Nets are provided in the supplementary material.}
\label{capacity}
\resizebox{10.5cm}{1.45cm}{
\begin{tabular}{lccc}
\hline
\multirow{2}{*}{Model} & original       & re-implementation & IBN-Net-a              \\ \cline{2-4} 
                       & top1/top5 err. & top1/top5 err.    & top1/top5 err.         \\ \hline
DenseNet121~\cite{huang2017densely}     & 25.0/-       & 24.96/7.85        & 24.47/7.25 (0.49/0.60) \\
DenseNet169~\cite{huang2017densely}     & 23.6/-       & 24.02/7.06        & 23.25/6.51 (0.79/0.55) \\
ResNet50~\cite{he2016deep}              & 24.7/7.8       & 24.27/7.08        & 22.54/6.32 (1.73/0.76)  \\
ResNet101~\cite{he2016deep}             & 23.6/7.1       & 22.48/6.23        & 21.39/5.59 (1.09/0.64) \\
ResNeXt101~\cite{xie2017aggregated}            & 21.2/5.6       & 21.31/5.74        & 20.88/5.42 (0.43/0.32) \\
SE-ResNet101~\cite{hu2017squeeze}          & 22.38/6.07     & 21.68/5.88        & 21.25/5.51 (0.43/0.37) \\ \hline
\end{tabular}
}
\end{center}
\end{table}
\setlength{\tabcolsep}{1.4pt}

\setlength{\tabcolsep}{4pt}
\begin{table}[!t]
\begin{center}
\caption{Results of IBN-Net variants on ImageNet validation set and Monet style set.}
\label{variants}
\resizebox{8.3cm}{1.6cm}{
\begin{tabular}{lcc}
\hline
\multirow{2}{*}{Model} & origin              & Monet                              \\ \cline{2-3} 
                       & top1/top5 err.      & top1/top5 err.                     \\ \hline
ResNet50               & 24.26/7.08          & 54.51/29.32 (30.24/22.24)          \\
IBN-Net50-a            & \textbf{22.54/6.32} & 51.57/27.15 (29.03/20.83)          \\
IBN-Net50-b            & 23.64/6.86          & \textbf{50.45/25.22 (26.81/18.36)} \\
IBN-Net50-c            & 22.78/\textbf{6.32} & 51.83/27.09 (29.05/20.77)          \\
IBN-Net50-d            & 22.86/6.48          & 50.80/26.16 (27.94/19.68)          \\
IBN-Net50-a\&d         & 22.89/6.48          & 51.27/26.64 (28.38/20.16)          \\
IBN-Net50-a\(\times2\) & 22.81/6.46          & 51.95/26.98 (29.14/20.52)          \\ \hline
\end{tabular}
}
\end{center}
\end{table}
\setlength{\tabcolsep}{1.4pt}

\begin{table}[!t]
\parbox{.45\linewidth}{
\centering
\caption{Comparison of IBN-Net50-a with IN layers added to different amount of residual groups. }
\label{INposition}
\resizebox{6cm}{0.6cm}{
\begin{tabular}{c|ccccc}
\hline
Residual groups    & none  & 1     & 1-2   & 1-3   & 1-4   \\ \hline
top1 err. & 24.27 & 23.58 & 22.94 & \textbf{22.54} & 22.96 \\
top5 err. & 7.08  & 6.72  & 6.40  & \textbf{6.32}  & 6.49  \\ \hline
\end{tabular}
}
}
\qquad
\parbox{.45\linewidth}{
\centering
\caption{Effects of the ratio of IN channels in the IBN layers. 'full' denotes ResNet50 with all BN layers replaced by IN. }
\label{INratio}
\resizebox{6cm}{0.6cm}{
\begin{tabular}{l|cccccc}
\hline
IN ratio  & 0     & 0.25  & 0.5   & 0.75  & 1     & full  \\ \hline
top1 err. & 24.27 & \textbf{22.49} & 22.54 & 23.11 & 23.44 & 28.56 \\
top5 err. & 7.08  & 6.39  & \textbf{6.32}  & 6.57  & 6.94  & 9.83  \\ \hline
\end{tabular}
}
}
\end{table}

\textbf{Model Capacity.}
To demonstrate the stronger model capacity of IBN-Net over traditional CNNs, we compare its performance with a number of recently prevalent CNN architectures on the ImageNet validation set.
As Table~\ref{capacity} shows, IBN-Net achieves consistent improvement over these CNNs, indicating stronger model capacity.
Specifically, IBN-ResNet101 gives comparable or higher accuracy than ResNeXt101 and SE-ResNet101, which either requires more time consumption or introduces additional parameters.
Note that our method brings no additional parameters while only add marginal calculations during inference phase.
Our results show that, dropping out some mean and variance statistics in features helps the model to learn from images with high appearance diversity.

\textbf{IBN-Net variants.}
We further study some other variants of IBN-Net.
Table.~\ref{variants} shows results for IBN-Net variants described in the method section.
All our IBN-Net variants show better performance than the original ResNet50 and less performance drop under appearance transform.
Specifically, IBN-Net-c achieves similar performance as IBN-Net-a, providing an alternative feature combining approach.
The modeling and generalization capacity of IBN-Net-d lies in between IBN-Net a and b, which demonstrates that preserving some BN features help improve performance, but loses generalization meanwhile.
The combination of IBN-Net a and d makes little difference with d, showing that the effects of INs on the main path of ResNet would dominate, eliminating the effects of those on the residual path.
Finally, adding additional IBN layers to IBN-Net-a brings no good, a moderate amount of IN features would suffice.

\textbf{On the amount of IN and BN.}
Here we study IBN-Nets with different amount of IN layers added.
Table.\ref{INposition} gives performance of IBN-Net50-a with IN layers added to different amount of residual groups.
It can be seen that the performance is improved with more IN layers added to shallow layers, but decreased when IN layers are added to the last residual group. This indicates that IN in shallow layers help to improve modelling capacity, while in deep layers BN should be kept to preserve important content information.
Furthermore, we study the effects of IN-BN ratio on the performance, as shown in Table.\ref{INratio}.
Again, the best performance is achieved at a moderate ratio 0.25-0.5, demonstrating the trade-off relationship between IN and BN.

\subsection{Cross Domain Experiments}

If models trained with synthetic data could be applied to the real world, it would save much effort for data collection and labelling.
In this section we study our model's capacity to generalize across real and synthetic domains on Cityscapes and GTA5 datasets.

\textbf{Cityscapes}~\cite{cordts2016cityscapes} is a traffic scene dataset collect from a number of European cities.
It contains high resolution \(2048 \times 1024\) images with pixel level annotations of 34 categories.
The dataset is divided into 2975 for training, 500 for validation, and 1525 for testing.

\textbf{GTA5}~\cite{Richter_2016_ECCV} is a similar street view dataset generated semi-automatically from the realistic computer game Grand Theft Auto V (GTA5).
It has 12403 training images, 6382 validation images, and 6181 testing images of resolution \(1914 \times 1052\) and the labels have the same categories as in Cityscapes.

\textbf{Implementation.} 
During training, we use random scale, aspect ratio and mirror for data augmentation.
We apply random crop on full resolution images for Cityscapes and \(1024 \times 563\) resized images for GTA5, because this leads to better performance for both datasets.
We use the "poly" learning rate policy with base learning rate set to 0.01 and power set to 0.9.
We train the models for 80 epochs.
Batch size, momentum and weight decay are set to 16, 0.9, and 0.0001 respectively.
When training on GTA5, we use a quarter of the train data so that the data scale matches that of Cityscapes.

As in \cite{chen2017deeplab}, we use ResNet50 with atrous convolution strategy as our baseline, and our IBN-Net follows the same modification.
We train the models on each dataset and evaluate on both, the results are given in Table~\ref{csGTA}.

\setlength{\tabcolsep}{4pt}
\begin{table}[!t]
\begin{center}
\caption{Results on Cityscapes-GTA dataset. Mean IoU for both within domain evaluation and cross domain evaluation is reported.}
\label{csGTA}
\resizebox{8.3cm}{2.1cm}{
\begin{tabular}{|l|l|lcc|}
\hline
Train                       & Test                        & Model       & mIoU(\%)      & Pixel Acc.(\%) \\ \hline
\multirow{6}{*}{Cityscapes} & \multirow{3}{*}{Cityscapes} & ResNet50    & 64.5          & 93.4           \\
                            &                             & IBN-Net50-a & \textbf{69.1} & \textbf{94.4}  \\
                            &                             & IBN-Net50-b & 67.0          & 94.3           \\ \cline{2-5} 
                            & \multirow{3}{*}{GTA5}       & ResNet50    & 29.4          & 71.9           \\ 
                            &                             & IBN-Net50-a & 32.5          & 71.4           \\
                            &                             & IBN-Net50-b & \textbf{37.9} & \textbf{78.8}  \\ \hline
\multirow{6}{*}{GTA5}       & \multirow{3}{*}{GTA5}       & ResNet50    & 61.0          & 91.5           \\ 
                            &                             & IBN-Net50-a & \textbf{64.8} & \textbf{92.5}  \\
                            &                             & IBN-Net50-b & 64.2          & 92.4  \\ \cline{2-5} 
                            & \multirow{3}{*}{Cityscapes} & ResNet50    & 22.2          & 53.5           \\ 
                            &                             & IBN-Net50-a & 26.0          & 60.9           \\ 
                            &                             & IBN-Net50-b & \textbf{29.6} & \textbf{66.8}  \\ \hline
\end{tabular}
}
\end{center}
\end{table}
\setlength{\tabcolsep}{1.4pt}

\begin{table}[!t]
\centering
\caption{Comparison with domain adaptation methods. Note that our method does not use target data to help adaptation. }
\label{DA}
\resizebox{7.1cm}{1.55cm}{
\begin{tabular}{|l|c|c|c|}
\hline
Method                   & mIoU & mIoU gain                     & Target data                    \\ \hline
Source only~\cite{hoffman2016fcns}              & 21.2 & \multirow{2}{*}{5.9}          & \multirow{2}{*}{w/}            \\ 
FCN wild~\cite{hoffman2016fcns}                 & 27.1 &                               &                                \\ \hline
Source only~\cite{zhang2017curriculum}              & 22.3 & \multirow{2}{*}{6.6}          & \multirow{2}{*}{w/}            \\ \
Curr. DA~\cite{zhang2017curriculum}                 & 28.9 &                               &                                \\ \hline
Source only~\cite{sankaranarayanan2017unsupervised}              & 29.6 & \multirow{2}{*}{\textbf{7.5}}          & \multirow{2}{*}{w/}            \\ 
GAN DA~\cite{sankaranarayanan2017unsupervised}                   & 37.1 &                               &                                \\ \hline
Ours - Source only       & 22.17 & \multirow{2}{*}{\textbf{7.5}} & \multirow{2}{*}{\textbf{w/o}} \\
Ours - IBN - Source only & 29.64 &                               &                                \\ \hline
\end{tabular}
}
\end{table}

\begin{table}[!t]
\centering
\caption{Finetune with different data percent.}
\label{finetune}
\resizebox{6.0cm}{0.65cm}{
\begin{tabular}{lcccc}
\hline
Data for finetune (\%) & 10   & 20   & 30            & 100   \\ \hline
ResNet50               & 52.7 & 54.2 & 58.7          & 63.84 \\ \hline
IBN-Net50-a            & 56.5 & 60.5 & 65.5          & 68.78     \\ \hline
\end{tabular}
}
\end{table}

\textbf{Results.} 
Our results are consistent with those on the ImageNet dataset.
IBN-Net shows both stronger modeling capacity within one dataset and better generalization across datasets of different domains.
Specifically, IBN-Net-a shows stronger model capacity, outperforming ResNet50 by 4.6\% and 3.8\% on the two datasets.
And IBN-Net-b's generalization is better, as the cross evaluation performance is increased by 8.5\% from Cityscapes to GTA5 and 7.5\% for the opposite direction.

\textbf{Comparison with domain adaptation methods.} It should be mentioned that our method is under the different setting with the domain adaptation works.
Domain adaptation is target domain oriented and requires target domain data during training, while our method does not.
Despite so, we show that the performance gain of our method is comparable with those of domain adaptation methods, as Table.~\ref{DA} shows.
Our approach takes an important step towards more generalizable models since we introduce built-in appearance invariance to the model instead of forcing it to fit into a specific data domain.

\textbf{Finetune on Cityscapes.} Another commonly used approach to apply a model on new data domain is to finetune it with a small amount of target domain annotations.
Here we show that with our more generalizable model, the data required for finetuning could be significantly reduced.
We finetune the models pretrained on the GTA5 dataset with different amount of Cityscapes data and labels.
The initial learning rate and the number of epochs is set to 0.003 and 80 respectively.
As Table.~\ref{finetune} shows, with only 30\% of Cityscapes training data, IBN-Net50-a outperforms resnet50 finetuned on all the data.

\subsection{Feature Divergence Analysis}

In order to understand how IBN-Net achieves better generalization, we analyse the feature divergence caused by domain bias in this section.
Similar to ~\cite{li2016revisiting}, our metric for feature divergence is as follows. 
For the output feature of a certain layer in a CNN, we denote the mean value of a channel as \(F\), which basically describes how much this channel is activated.
We assume a Gaussian distribution of \(F\), with mean \(\mu\) and variance \(\sigma^2\).
Then the symmetric KL divergence of this channel between domain A and B would be:
\begin{align}
  D(F_{A} || F_{B}) &= KL(F_{A} || F_{B}) + KL(F_{B} || F_{A}) \\
  KL(F_{A} || F_{B}) &= log\frac{\sigma_{B}}{\sigma_A} + \frac{\sigma_{A}^{2} + (\mu_A - \mu_B)^2}{2\sigma_{B}^{2}} - \frac{1}{2}
\end{align}
Denote \(D(F_{iA}||F_{iB})\) as the symmetric KL divergence of the \(i\)th channel, then the average feature divergence of the layer would be:
\begin{align}
  D(L_{A} || L_{B}) &= \frac{1}{C}\sum_{i=1}^{C}D(F_{iA}||F_{iB})
\end{align}
where \(C\) is the number of channels in this layer. This metric provides a measurement of the distance between feature distribution for domain A and that for domain B.

\begin{figure*}[!t]
\centering
\includegraphics[width=7.0cm]{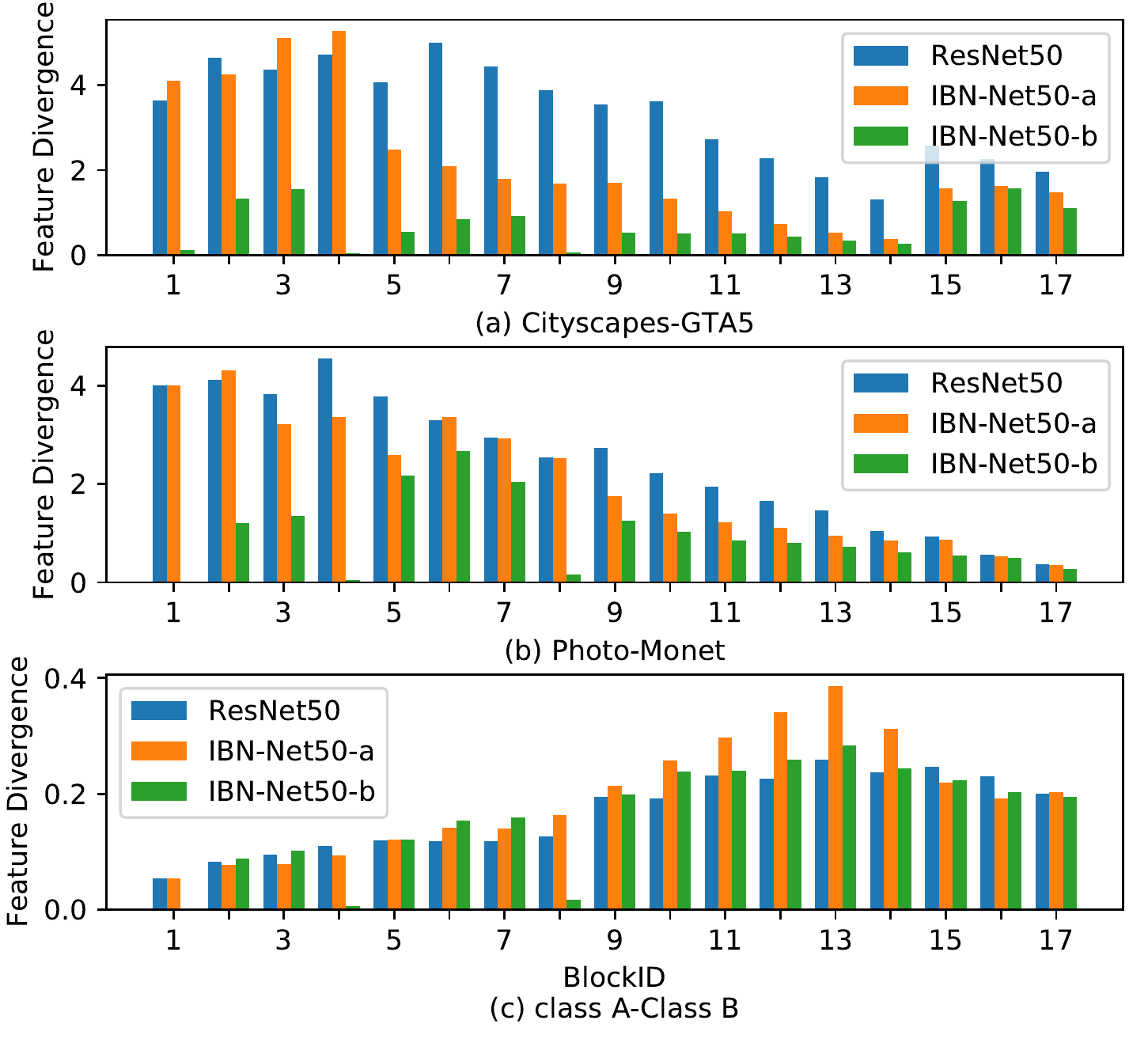}
\caption{\label{FDA} Feature divergence caused by (a) real-virtual appearance gap, (b) style gap, (c) object class difference. }
\end{figure*}

To capture the effects of instance normalization on appearance information and content information, here we consider three groups of domains.
The first two groups are "Cityscapes-GTA5" and "photo-Monet", which differs in complex appearance.
To build two domains with different contents, we split the ImageNet-1k validation set into two parts, with the first part containing images with 500 object categories and the second part containing those with the other 500 categories.
Then we calculate the feature divergence of the 17 ReLU layers on the main path of ResNet50 and IBN-Net50.
The results are shown in Fig.~\ref{FDA}.

It can be seen from Fig.~\ref{FDA}(a)(b) that in our IBN-Net, the feature divergence caused by appearance difference is significantly reduced.
For IBN-Net-a the divergence decreases moderately while for IBN-Net-b it encounters sudden drop after IN layer at position 2,4,8.
And this effect lasts till deep layers where IN is not added, which implies that the variance encoding appearance is reduced in deep features, so that their interference with classification is reduced.
On the other hand, the feature divergence caused by content difference does not drop in IBN-Net, as Fig.~\ref{FDA}(c) shows,
showing that the content information in features are well preserved in BN layers.

\textbf{Discussions.} These results give us an intuition of how IBN-Net gains stronger generalization.
By introducing IN layers to CNNs in a clever and moderate way, they could work in a manner that helps to filter out the appearance variance within features.
In this way the models' robustness to appearance transforms is improved, as shown in our experiments.

Note that generalization and modelling capacity are not uncorrelated properties.
On one hand, intuitively appearance invariance could also help the model to better adapt to the training data of high appearance diversity and extract their common aspects.
On the other hand, even within one dataset, appearance gap exists between the training and testing set, in which case stronger generalization would also improve performance.
These could be the reasons for the stronger modelling capacity of IBN-Net.

\section{Conclusions}

In this work we propose IBN-Net, which carefully unifies instance normalization and batch normalization layers in a single deep network to increase both modeling and generalization capacity.
We show that IBN-Net achieves consistent improvement over a number of classic CNNs including VGG, ResNet, ResNeXt, and SENet on ImageNet dataset.
Moreover, the built-in appearance invariance introduced by IN helps our model to generalize across image domains even without the use of target domain data.
Our work concludes the role of IN and BN layers in CNNs: IN introduces appearance invariance and improves generalization while BN preserves content information in discriminative features.

~\\
\textbf{Acknowledgement.} This work is partially supported by SenseTime Group Limited, the Hong Kong Innovation and Technology Support Programme, and the National Natural Science Foundation of China (61503366).

%
%
%
%

\bibliographystyle{splncs04}
\bibliography{egbib}
\end{document}